\def\year{2021}\relax
\documentclass[letterpaper]{article} 
\usepackage{aaai21}  
\usepackage{times}  
\usepackage{helvet} 
\usepackage{courier}  
\usepackage[hyphens]{url}  
\usepackage{graphicx} 
\usepackage{natbib}  
\usepackage{caption} 
\usepackage{hyperref}       
\usepackage{booktabs}       
\usepackage{amsmath}
\usepackage{algorithm}
\usepackage{algorithmic}
\usepackage{pdfpages}
\title{Automated Model Compression by Jointly Applied Pruning and Quantization}

\begin{document}
\author 
{

        Wenting Tang,\textsuperscript{\rm 1}
        Xingxing Wei, \textsuperscript{\rm 1}
        Bo Li \textsuperscript{\rm 1} \\
}
\affiliations {
    \textsuperscript{\rm 1} Beijing Key Laboratory of Digital Media, 
   School of Computer Science and Engineering, 
   Beihang University
   Beijing, China \\
    wtang13@buaa.edu.cn, xxwei@buaa.edu.cn, boli@buaa.edu.cn
}
\maketitle

\begin{abstract}
  Although deep neural networks (DNNs) achieve excellent performance
  in real-world computer vision tasks,
  network compression may be necessary to
  adapt DNNs into edge devices such as mobile phones.
  In the traditional \emph{deep compression} framework,
  iteratively performing network pruning and quantization can reduce the model size and computation cost to meet the deployment requirements. 
   However, such a step-wise application of pruning and
    quantization may lead to suboptimal solutions and unnecessary  time consumption.
    In this paper, we tackle this issue by integrating network pruning and quantization
    as a unified \emph{joint compression} problem, and then use AutoML to automatically solve it.
    We find the pruning process can be regarded as the channel-wise quantization with 0 bit. Thus, the separate two-step pruning and quantization can be simplified as the one-step quantization with mixed precision.  This unification not only simplifies the compression pipeline but also avoids the compression divergence.
        To implement this idea, we propose the \textbf{A}utomated model compression by \textbf{J}ointly applied \textbf{P}runing and \textbf{Q}uantization (AJPQ). AJPQ is designed with a hierarchical architecture: the layer controller controls the layer sparsity and the channel controller decides the bitwidth for each kernel. Following the same importance criterion, the layer controller and the channel controller collaboratively decide the compression strategy. With the help of reinforcement learning, our one-step compression is automatically achieved. Compared with the state-of-the-art automated compression methods, our method obtains a better accuracy while reducing the storage considerably. For fixed precision quantization, AJPQ can reduce more than $5\times$ model size and $2\times$ computation with a slight performance increase for Skynet in the remote sensing object detection. When mixed precision is allowed, AJPQ can reduce $5\times$ model size with only ${1.06\%}$ top-5 accuracy decline for MobileNet in the classification task.

\end{abstract}

\section{Introduction}\label{section:intor}

Although deep neural networks (DNNs) achieve excellent performance
in real-world computer vision tasks,
  network compression may be necessary to adapt DNNs into
  edge devices such as mobile phones and surveillance camera \cite{zou2019object}.
   To obtain a balance between the model's performance and reduction,
   different approaches of model compression are proposed
   such as knowledge distillation, network pruning and binary
convolutional network \cite{hinton2015distilling, han2015learning, rastegari2016xnor}, etc.

Either network pruning or network quantization can achieve model compression
by reducing the size or bitwidth of weights.
 However, those compression techniques sometimes can be
 heavy-duty of human work requiring a tedious parameter adjusting process.
 Although some criterions \cite{li2017pruning, han2015learning}
 can help to identify redundant connections during pruning, it may involve manual work to decide proper layer sparsity. Likewise, when mixed precision \cite{elthakeb2018releq} is introduced to fully leverage advantages of the cutting-edge hardware, the appropriate combination of each layer's or kernel's bitwidth needs to be manually explored. To address this issue, the AutoML technique is introduced to perform automated compression \cite{he2018amc,manessi2018automated,wang2019haq,gao2019dynamic}. In these methods, the \emph{deep compression} framework proposed by \citet{han2016deep} contains a three stage pipeline:
pruning, quantization and Huffman coding to reduce the storage step by step. To further improve the performance for each step, they utilize the AutoML(Automated Machine Learning) to perform automated pruning \cite{he2018amc} and automated  quantization \cite{wang2019haq}, respectively. 
\begin{figure}[t]
  \centering
  \includegraphics[width=1.0\linewidth]{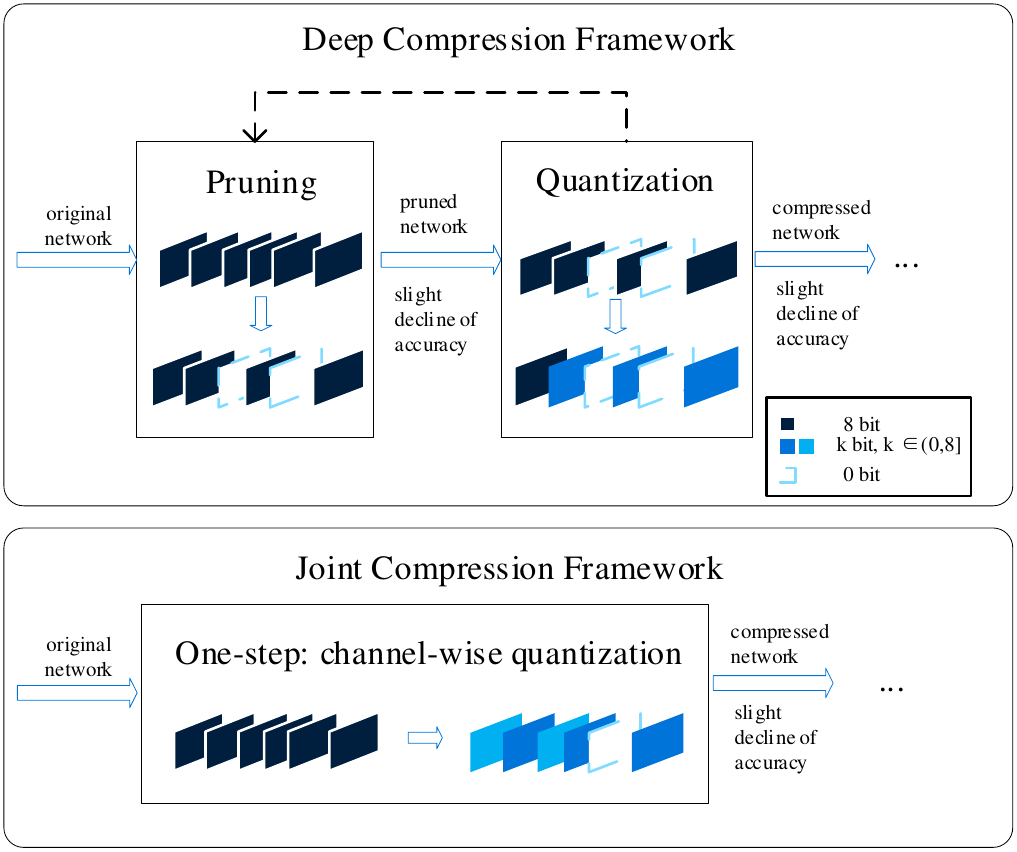}
  \caption{The \emph{deep compression} \cite{han2016deep}
  and our \emph{joint compression}. The \emph{deep compression} framework applies network pruning and quantization step by step, which means network quantization only starts when pruning finish. Based on compression requirements, our \emph{joint compression} framework can spontaneously switch among different compression techniques for different layers.}
  \label{fig1}
  \vspace{-.4cm}
\end{figure}

 Figure \ref{fig1} illustrates the framework of the \emph{deep compression}. In the pruning phase, deep compression takes the original network as input, and then removes the
 unnecessary connections, outputting a pruned model. In the quantization phase, deep compression takes the pruned network as input, and then produces
 a nearly lossless compressed model represented by the integer weights. In each phase, the optimal compression can be achieved because the AutoML technique can search the most available compression strategy according to hardware  constraint. However, in the view of the whole compression, this simple combination of pruning and quantization is suboptimal. The pruning and quantization should interact with each other to achieve the global optimal balance between the performance holding and computation reduction, rather than their own optimum. 
 What's more, because two optimization problems need solving, this step-wise design is time consuming.
 

To tackle this issue, we incorporate the pruning and quantization into a \emph{joint compression} framework, and then use AutoML to jointly search the globally optimal compression strategy. Specifically, 
we find the pruning process can be regarded as the channel-wise quantization with 0 bit. Thus, the separate two-step pruning and quantization can be simplified as the one-step quantization with mixed precision. 
Thanks to this simplification, our framework has the potential to provide a better compression solution than the
\emph{deep compression} framework. A challenge for our framework is how to perform the channel-wise quantization with mixed precision from 0 to $N$ bits. Because each channel has many candidate bits, the searching space is large. For this reason, the current automated quantization methods usually perform mixed precision across different layers \cite{wang2019haq, elthakeb2018releq} (layer number is much less than channel number). In this paper, we design a novel hierarchical architecture: the layer controller controls the layer sparsity
and the channel controller decides the bitwidth for each channel.
 Following the same importance criterion, the layer controller and channel controller collaboratively decide a quantization plan for the whole network. In this way, we solve the inefficiency problem. 

\begin{figure}[t]
  \centering
  \includegraphics[width=1.0\linewidth]{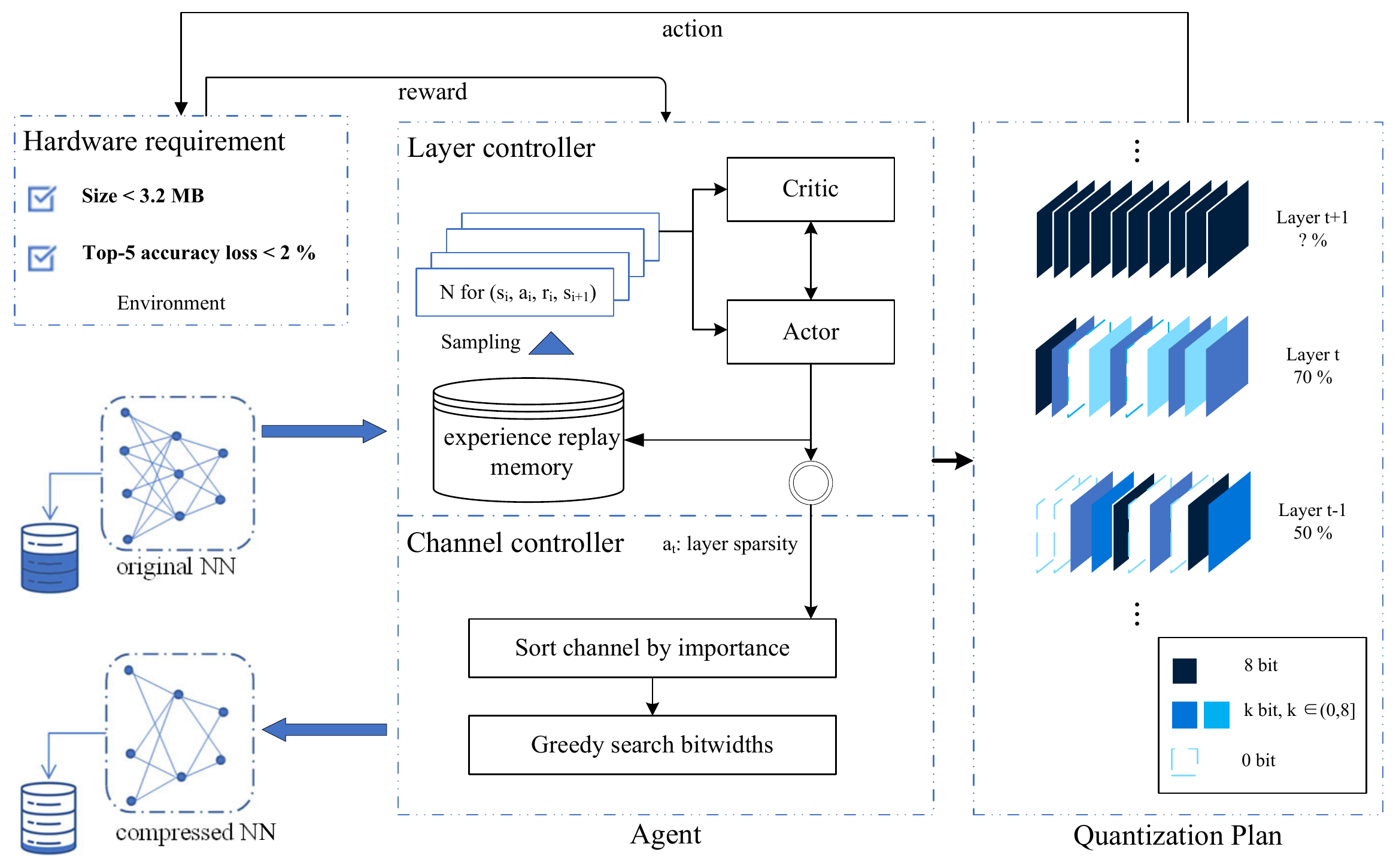}
  \caption{Overview of the Automated Model Compression by Jointly applied
  Pruning and Quantization (AJPQ) engine. For specific compression requirements and a model, AJPQ searches the quantization plan layer by layer. At each layer, the layer controller decides the layer sparsity, and then the kernel controller determines the bit -width for preserved channels. After searching the quantization plan for specific episodes, AJPQ outputs a compressed model by the quantization plan with the highest reward.}
  \label{fig2}
\end{figure}

Technically speaking, we propose Automated Model Compression by Jointly applied Pruning and Quantization (AJPQ)
as shown in Figure \ref{fig2}.
Inspired by layer sparsity control in AMC \cite{he2018amc},
AJPQ's layer controller can automatically search the optimal preserved ratio in continuous space for each layer by DDPG(Deep Deterministic Policy Gradient) \cite{silver2014deterministic}.
With reduction achieved by removing trivial channels, the channel controller only needs to decide the non zero bitwidth for the rest of the channels, whose number is reduced.
By using size constraint as the upper bound and applying weight importance to compute the lower bound, greedy searching is capable of quickly generating a quantization
plan for each layer. For this plan,
 the environment (hardware) will respond with the state information and reward to the agent to help correct the quantization plan in the next iteration.

In summary, our contributions can be concluded in the three aspects:
\begin{itemize}
\item [1)]
A \emph{joint compression} framework is proposed
to accelerate the searching for a globally optimal compression strategy while retaining the performance of the compressed model.
\item [2)]
We propose a new viewpoint that pruning process can be regarded as the channel-wise quantization with 0 bit, and furthermore, gives a hierarchical architecture to efficiently and automatically solve the channel-wise quantization with mixed precision.
\item [3)]
We evaluate AJPQ approach with the state-of-the-art
automated compression tools AMC \cite{he2018amc} and HAQ \cite{wang2019haq} on MobileNet \cite{howard2017mobilenets} and Skynet.
A better performance with less time consumption is achieved.
What's more, AJPQ can reduce more than $5\times$ model size and $2\times$ computation while achieving the performance with $4.5\%$ improvement for Skynet in the remote sensing object detection.

\end{itemize}

The following sections are orginzed as follow: previous work in network compression is summarized in \nameref{section:rw}; detailed design of AJPQ is presented in \nameref{section:method}; effectiveness of AJPQ for the classification and detection is presented in \nameref{section:exp}; conclusion and further improvement is in \nameref{sec:con}.

\section{Related Work}\label{section:rw}
As network pruning can be defined as a channel selection problem
followed by a specific criterion,
extensive works focus on evaluating the redundancy or
importance of channels. For example,
the weight values of connections are initially used to evaluate the importance of
connections in a trained model \cite{li2017pruning, han2015learning, he2019filter, zhuang2018discrimination, ye2018rethinking}.
In contrast to directly utilizing trained weights,
\citet{he2019filter} represent the common information by geometric median among filters
as an attempt to overcome the "small norm deviation" and "large minimum norm" problems brought by the norm-based criterion \cite{li2017pruning}.
\citet{anwar2017structured} identify the trivial and redundant
 connection by a genetic algorithm to adapt to the different pruning particle.
 Although these criteria are efficient in computation and application,
 defining preserve ratio or sparsity for each layer is manual work.
 
 To get rid of the tedious process of parameter adjustment,
 \cite{zhuang2018discrimination, ye2018rethinking, liu2019metapruning, luo2017thinet, zhu2018to, gao2019dynamic}
 train a pruned network from the scratch. For example, 
 \citet{zhuang2018discrimination} augment the meaning of channel importance with the discriminative power.
 \citet{ye2018rethinking} form the channel selection as an optimization problem and solve it by ISTA(Iterative Shrinkage-Thresholding Algorithm) in batch normalization applied DNNs.
 Inspired by NAS \cite{zoph2017neural}, \cite{liu2019metapruning}
 keeps the model with the best performance from a set of random generated pruned models.

As the pruning process can be defined as a Markov decision process \cite{bertsekas1995dynamic}, techniques in AutoML(Automated Machine Learning) domain are applied to efficiently search the compression plan
in a layer-by-layer manner \cite{elthakeb2018releq,he2018amc, wang2019haq, ashok2018n2n}.
 To support deep quantization which allows mixed precision in network representation,
 HAQ and ReLeQ \cite{wang2019haq, elthakeb2018releq} are proposed while ReLeQ \cite{elthakeb2018releq}
 requires a fine-tuning stage beside the reinforcement learning searching loop.
  Those reinforcement learning approaches enable more controls
  during pruning or quantization related schedule process.

After \emph{deep compression} framework \cite{han2016deep},
plenty of works are inspired \cite{manessi2018automated, he2018amc,chen2017eyeriss, wen2016learning,zhu2018to}. For example, \cite{manessi2018automated}, \cite{he2018amc} and \cite{zhu2018to} aim to automatically prune connections.
 While \citet{manessi2018automated} improve the threshold-based channel selection in \cite{han2016deep}.
  \cite{zhu2018to} and \cite{he2018amc} schedule preserve ratio for each layer by a sparsity function and RL agent respectively.
  To achieve a considerable reduction in convolutional layers of the pruning process in deep compression,
   \cite{wen2016learning} proposes to learn the structured sparsity.
   Utilizing the \emph{deep compression} framework, \cite{chen2017eyeriss} deploys an energy-efficient model for real-world application.
 
   There are also explorations to achieve one-step compression without following the \emph{deep compression} framework such as NAS(Neural architecture search) \cite{zoph2017neural} and binary convolutional
   network \cite{rastegari2016xnor}. Considering the high computations resource requirements, NAS may not be an economical solution for general usage.
   The binary convolutional network may be the answer, but it cannot replace all existing useful DNNs so far.
 
   Overall, AJPQ automatically achieves the pruning and quantization in the one-step compression, which is different from the above works.

\begin{figure}[htp]
  \centering
  \includegraphics[width= 1.0\linewidth ]{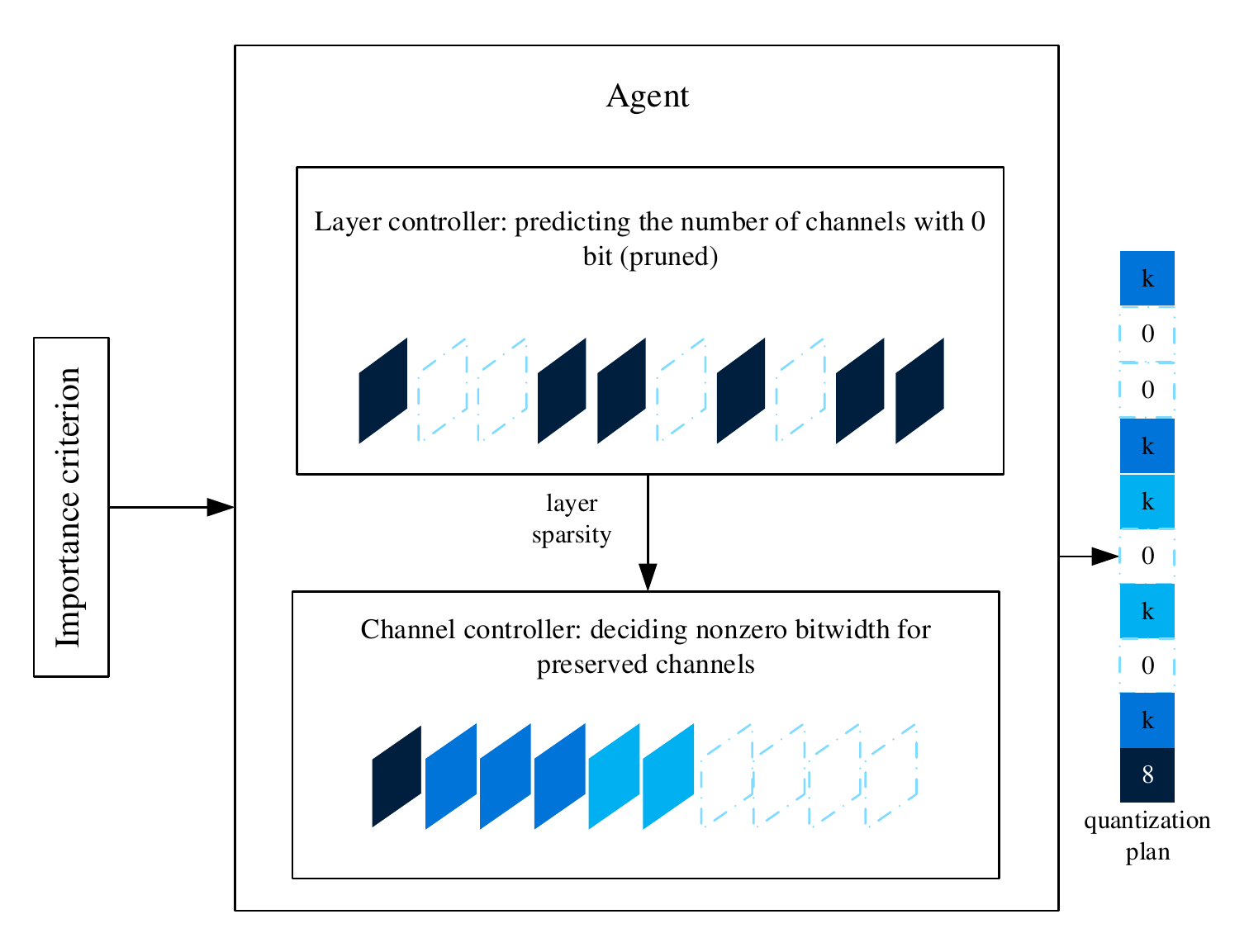}
  \caption{Roles for layer and channel controller in our method. Guided by the same importance criterion at each layer, the agent generates a quantization plan to deliver positive reward with respect to the performance and compression. The quantization plan is generated by the layer controller and the channel controller together: the layer controller decides channels with zero bit-width; the controller decides non-zero bit-width for the preserved channels.}
  \label{fig3}
\end{figure}
\section{Methodology}\label{section:method}
    Following the \emph{joint compression} framework,
    the AJPQ is proposed for searching the quantization plan.
    The layer-wise searching for a proper quantization plan can be formulated as
    a dynamic programming problem. Considering a DNN model
    with $n$ quantizable layers and the channel number for each layer is represented by $I^l,l=1,...,n$, thus $\mathcal{I} = \{ I_1,..., I_n\}$, our goal is to balance the performance and storage by assigning each channel with the proper bitwith. Let us denote that $s_l$ is an environment state, $P^l$ is a channel-wise quantization plan, $r_l$ is the reward of the generated channel-wise quantization plan.We assume that $r_0,r_1,...,r_{n-1}$ are independent. Thus, the quantization state changes as follows:
    \begin{equation}
      s_{l+1} = f(s_l, P^l, r_l)
    \end{equation}
    For the $l$-th layer, the problem space is $(bit_{max}+1)^{I_l}$
    with allowed the bitwidth in [0, $bit_{max}$], where the $bit_{max}$ is the maximal allowed bitwidth and is set as 8 in default.
    Because $(bit_{max}+1)^{I_l}$ is very large, the amount of the possible permutations of $P^l$ is massive. The previous work has shown that eliminating the redundant channels in a trained model is nearly lossless \cite{he2019filter}, thus, removing the trivial channels will make the problem space shrink. A widely used metric to measure the channel's redundancy is the importance score $\{imp_i^l|i=1,...,I_l\}$, which is computed by the $L1$-norm of the corresponding kernel values in the $l$-th layer \cite{li2017pruning}. For the preserved channels, we narrow the searching space by the assumption that the more important a channel is, the longer bitwidth it should hold. Therefore, by using the identical importance score $imp_i^l$, the minimal bitwidth $minbit^l_c$ is computed for the $c$-th preserved channel. $minbit^l_c$ is used as the lower bound in the searching, and the upper bound is  $bit_{max}$. So the problem space is shrunk to $\prod_{c=1}^{c_{nz}^l}(bit_{max}-minbit^l_c$, where $c_{nz}^l$ is the number of nonzero bitwith channels. With the approaching goal and a narrowed searching space, the non-zero mixed bitwidths $\mathcal{B} = \{b_1, ...,b_{c_{nz}^l}\}$ for the preserved channels should be solved by a greedy algorithm without dramatic performance decline. In this way, this dynamic programming problem
    is efficiently optimized.
    
    Specifically, the layer controller in our method identifies the redundancy, and predicts the layer sparsity by DDPG \cite{silver2014deterministic}.
    The channel controller performs channel selection and decides the different bitwidths for the nontrivial channels by a greedy algorithm.
\begin{algorithm}[hb!]
\caption{Produce quantization plan for the whole network}
\label{alg1}
\hspace*{0.02in} {\bf Input :} $\mathcal{N}$, $\mathcal{N}_{size}$, $n$, $\mathcal{I}$, $\mathcal{N'}_{size}$, each layer's size $S$, total searching episodes $e$, required model compression ratio $sc$, upper bound of the bitwidth $bit_{max}$;
\begin{algorithmic}[1]
\STATE Initialize $s_{0}$ represented by Equation (\ref{e0}) and the current searching episodes $e_{cur}=0$
\WHILE{$e_{cur}<e$}
    \STATE $l=0$, $P=\{bit_{max}\}$
    \STATE Initialize the critic network $\mu$, the actor network $\theta$, exploration noise $\sigma = 0.9$, each quantizable layer size $S=\{S^l|l=1,...,n\}$ by original model
    $\mathcal{N}$ 
    \WHILE{$l < n$}
        \STATE $a_l = TN(\mu(s_l|\theta_{l}^{\mu}),\sigma ^{2},0,1))$
        \STATE Compute the number of nonzero bit channels $c_{nz}^l = min(I^l, max(1, \lceil I^l \times a_l\rceil))$
        \STATE Compute $imp^l_i=\left | \sum_{i=1}^{i=I^{l+1}} (F^l_i)\right |$ by the value of the according filters $F^l_i$
        \STATE Sort $imp_i^l$ in the descending order 
        \STATE Compute index of the preserved channels $I_{nz}^l$ = $top_{c_{nz}^l}\{imp_i^l|,i=1,...,I_l\}$
        \STATE For $i\notin I_{nz}^l$, $P^l_i=0, minbit^l_i=0$
        \STATE For $i\in I_{nz}^l$, $P^l_i=bit_{max}$, $minbit^l_i=bit_{max} \times\frac{imp^l_i}{\max\limits_{c}(imp^l_c)}$
        \STATE Update $S^l$, $\mathcal{N}_{size}$ by quantization plan $P^l$ 
        \IF{$l$ == $n-1$}
            \WHILE{$\mathcal{N'}_{size} > sc\times\mathcal{N}_{size}$}
                \STATE $l'=\max\limits_{i \in {1,...,n}}(S^i) $
                \STATE $c'=\min\limits_{j \in I_{nz}^l}(imp_j^l)$
                \IF{$P^{l'}_{c'} > minbit^{l'}_{c'}$}
                    \STATE $P^{l'}_{c'} = P^{l'}_{c'} - 1$
                \ENDIF
                \STATE Update $S^l$, $\mathcal{N}_{size}$ by quantization plan $P^l$
            \ENDWHILE
        \ENDIF
        \STATE Update reward $r_l$ by Equation (\ref{e3}) and $S_l$
        \STATE $l$ = $l$ + 1 
    \ENDWHILE
    \STATE Update layer controller with $a_l$, $s_l$ by DDPG \cite{silver2014deterministic}
    \STATE $e_{cur}$ = $e_{cur}$ + 1
\ENDWHILE

\end{algorithmic}
\end{algorithm}
\subsection{Layer Controller}
\label{3.2.1}
Reinforcement learning is deployed for identifying layer sparsity $a_l$, which means the ratio between the number of nonzero bit channels and the total amount of original channels.
As we define that $a_l \in (0,1]$, the DDPG algorithm is used to solve the continuous action space of $a_l$.

\textbf{State Space}: The state space $\mathcal{M}=\{s_l|l=1,...,n\}$ is a finite set consisting of all quantizable layers for the model to be compressed.
The state $s_l$ is characterized by the following features:
\begin{equation} \label{e0}
\begin{split}
    (idx,t,c_o^l,c_i^l,w,h,stride,k,reducedFLOPs,\\
    restFLOPs,reducedSize,restSize,a_{l-1})
\end{split}
\end{equation}
where for a quantizable layer, $idx$ is the index, $t$ is the type of layer (fully connected layer or convolutional layer).
$c_o^l$ is the number of output channels and $c_i^l$ is the number of input channels.
$w$, $h$ are width and height for each input feature map.
$stride$, $k$ are the value of the stride and kernel size which will be 0 and 1 for fully connected layer, respectively.
$reducedFLOPs$ and $restFLOPs$ are the numbers of the reduced and the rest FLOPs
for the current quantization plan;
$reducedSize$ and $restSize$ are the amounts of the reduced and the rest storage for the current quantization plan.
The state is normalized at each update because it is friendly for actor and critic network \cite{ashok2018n2n}. As agent explores in a layer-by-layer manner, such features are required to seek a proper quantization plan.

\textbf{Reward}: The reward is used to evaluate the captured trade-off by comparing the performance
and the model size between the compressed model $\mathcal{N'}$ and the original model $\mathcal{N}$.
For $\mathcal{N}$ on classification task, the top-5 accuracy is $\mathcal{N}_{acc}$; for $\mathcal{N}$ on detection task, the average precision(AP) with IoU(intersection over union)=0.3 is $\mathcal{N}_{AP}$. $\mathcal{N}_{size}$ is the original model size. Our goal is compressing model
$\mathcal{N}$ within the $s\times\mathcal{N}_{size}$, where $s$ is decided by the hardware constraints and $s\in(0,1]$. The reward $r$ is derived by:
\begin{equation}
  r = \left\{\begin{matrix}
 -10.0& , \mathcal{N'}_{size} > s \times \mathcal{N}_{size}\\
 0.1 \times (\mathcal{N}'_{acc} - \mathcal{N}_{acc})& , \mathcal{N'}_{size} \leq s \times \mathcal{N}_{size}\\
 \mathcal{N}'_{AP} - \mathcal{N}_{AP}&, \mathcal{N'}_{size} \leq s \times \mathcal{N}_{size}
\end{matrix}\right.\label{e3}
\end{equation}

\begin{table*}[t]
\centering
\begin{tabular}{   l c c c  c   c }
\toprule
Methods                  & Accuracy       & Accuracy       & Compressed          & FLOPs                & Searching \\
$ $                      &(top-1)          &(top-5)         &ratio (bit)                     &ratio             &  episodes                  \\
\midrule
AMC+Fixed& 66.214($\downarrow$5.426)        & 85.330($\downarrow$4.14)          & 0.2484                      & \textbf{0.8005}                & 500                \\
AJPQ (Fixed)                & \textbf{67.020($\downarrow$4.62)}          & \textbf{87.110($\downarrow$2.36)}          & \textbf{0.214}                       & 0.817                 & 500  \\
\hline
AMC+HAQ (layer-wise)                 & \textbf{66.690($\downarrow$4.95)}        & 85.2($\downarrow$4.27)            & 0.207                       &\textbf{0.8005}                & 900       \\
AJPQ (layer-wise)  & 66.280($\downarrow$5.36)          & \textbf{87.00($\downarrow$2.47)}        & \textbf{0.195}                       & 0.808                 & \textbf{500}                \\
\hline
AJPQ (channel-wise) &\textbf{69.480}($\downarrow$\textbf{2.16})     &\textbf{88.410}($\downarrow$\textbf{1.06})      &\textbf{0.189}    &0.922   &\textbf{100}   \\\bottomrule
\end{tabular}
\caption{The accuracy, compressed ratio, and FLOPs ratio achieved by AMC \cite{he2018amc}, HAQ \cite{wang2019haq}, and our method on MobileNet. }
\label{tb1}
\end{table*}
\subsection{Channel Controller}
\label{3.2.2}
The channel controller is responsible for deciding the permutation of the bit-widths for the preserved channels. 
As shown in Algorithm \ref{alg1}, the number of preserved channels $c_{nz}^l$ is identified by layer sparsity $a_l$ at $l$-th layer. Then based on $c_{nz}^l$ and channels' importance $imp^l$, index of preserved channels $I_{nz}^l$ is identified. 
For greedy searching a quantization plan, the lower bound is represented by the
minimal bit-width $minbit$, and the upper bound is the maximal allowed bitwidth $bit_{max}$. When the compressed model size is larger than the
size constraint, the channel controller first selects a layer and then declines the bit-width from the trivial channels.
The channel controller is also compatible with the layer-wise quantization and the fixed precision quantization. Either the layer-wise quantization or the fixed precision is a special case of the channel-wise quantization. What's more, two layer selection rules are implemented in the AJPQ. The "large layer first" rule selects the layer with the largest layer size from all quantizable layers, the “deep layer first rule” adjusts the quantization plan from the deepest quantizable layer.

\subsection{Joint Decision}
\label{3.2.3}
As stated before, each layer's quantization plan is a joint decision made by the layer controller and the channel controller.
A hierarchical structure is designed to fulfill this target as shown in Figure \ref{fig3}.
An identical importance criterion guides the process of eliminating the redundancy and preserving channels.
After the layer controller deciding a layer sparsity,
the channel controller acts like a regulator: retaining the preserved channels with a high bit-width
when the layer sparsity is too high to maintain the performance and reducing some preserved
channels' bitwidth when the layer sparsity is too low to perform necessary compression. In this way, the elements associated with the trimmed channels' are set to 0 bit, while the other elements are retained in the proper bitwidth. Our method is concluded in Algorithm\ref{alg1}.

\section{Experimental Results}\label{section:exp}
\subsection{Experiment Setting}
 \textbf{Data set and Model:} For classification, 20,000 images of ImageNet\cite{deng2009imagenet} is used for state initialization and performance recovery; while 10,000 images of it is used for the validation. For detection, a remote sensing data-set containing bay area, coastline and ocean with 5702 images is used; While 1900 for validation and 3801 for state initialization and performance recovery. In the following experiments, we compare our method with the baselines on MobileNet\cite{howard2017mobilenets} for classification and Skynet\cite{zhang2020skynet} for detection. Skynet is a new hardware-efficient DNN specialized in object detection and tracking in aerial images. It only contains 6 depth-wise separable convolution\cite{howard2017mobilenets} blocks for feature extraction.\\
 \textbf{Metrics:} Top-1 and top-5 accuracy are used to estimate the performance of the compressed model in the classification task; AP (Average Precision) is used to evaluate the performance of the compressed model in the detection task.
 For the compressed model $\mathcal{N'}$ with the computation $\mathcal{N'}_c$ and  model size $\mathcal{N'}_s$, the original model $\mathcal{N}$ with the computation $\mathcal{N}_c$ and  model size $\mathcal{N}_s$, the compressed ratio is: $r_{comp} \,=\,\frac{\mathcal{N'}_s}{\mathcal{N}_s}$
 and the FLOPs ratio is: $r_{FLOPs}\,=\,\frac{\mathcal{N'}_c}{\mathcal{N}_c}$.
 The searching episode represents the time consumption for each method. \\
\textbf{Implementation Details} For AJPQ, the
maximal sparsity ratio $a_{max}$ for all quantizable layers is set to 1.0.
 The minimal sparsity ratio $a_{min}$ is set to 0.7 for the fully connected layers and 0.6 for convolutional layers. The original MobileNet and Skynet uses 32-bit floats, and the compressed models uses 8-bit integers. The batch size is set to 60 in the validation.

\subsection{Baselines}
 Here, different pruning and quantization methods are deployed into the \emph{deep compression} framework to form the baselines in our experiments.\\
\textbf{AMC+Fixed:} The AMC \cite{he2018amc} is used as the pruning component in the \emph{deep compression} framework. Fixed precision is achieved by the k-means \cite{macqueen1967some}. The bit-width of the compressed model is 8.\\
\textbf{AMC+HAQ:} To support the quantization with mixed precision, the HAQ \cite{wang2019haq} is used as the quantization component in the \emph{deep compression} framework. The HAQ can perform the mixed precision across different layers, and will automatically find the available bit-width for each layer. In this way, AMC+HAQ accomplish the quantization with mixed precision.

\begin{table*}[t]
\centering
\begin{tabular}{   l c c c    c }
\toprule
Methods                  & AP              & Compressed          & FLOPs                & Searching \\
$ $                      &                  &ratio (bit)         &ratio             &  episodes  \\
\midrule
AMC+Fixed&                 0.682($\downarrow$ 0.021)&                       0.242&                0.959&            150              \\
AJPQ (Fixed)&              \textbf{0.703}($\uparrow$\textbf{0.001}) &                     \textbf{0.197}&                 \textbf{0.744}&           150 \\
\hline
AMC+HAQ (layer-wise)&      0.728($\uparrow$0.025)&                     0.195&                 0.959&               450        \\
AJPQ (layer-wise)  &       \textbf{0.731}($\uparrow$\textbf{0.028})    & \textbf{0.188}        &\textbf{0.931}                      & \textbf{150}  \\
\hline
AJPQ (channel-wise) &      \textbf{0.747}($\uparrow$\textbf{0.045}) &      \textbf{0.1973}&   1.000&  300\\\bottomrule
\end{tabular}
\caption{The accuracy, compressed ratio, and FLOPs ratio achieved by AMC \cite{he2018amc}, HAQ \cite{wang2019haq}, and our method on SkyNet. }
\label{tb2}
\end{table*}

\subsection{Results and Analysis}

\textbf{Channel-wise Deep Quantization:}
  As shown in Table \ref{tb1},
   \emph{AJPQ (channel-wise)} achieves the highest size reduction 0.189 and the best model performance (88.41\% versus the top-5 accuracy) among all the methods with only 100 searching episodes.
  The slight decline in performance demonstrates that the importance criterion can guide channel-wise quantization to attain a proper trade-off among different compression techniques. After thorough exploration, the AJPQ achieves a noticeable detection performance increase by 0.045 only with quantization in Table \ref{tb2}. To maintain performance, the AJPQ wisely chooses the compression technique rather than simply combining the pruning and quantization.  

\textbf{Fixed precision:} As illustrated in Table \ref{tb1}, \emph{AJPQ (Fixed)} retains a 1.67\% higher top-5 accuracy with a nearly 0.03 larger size reduction
than the \emph{AMC+Fixed} approach. Under the same time consumption, the \emph{AJPQ (Fixed)} captures a better trade-off than the \emph{AMC+Fixed} approach.
This result illustrates that the \emph{deep compression} framework may be trapped into sub-optimal results by making compression decisions with local knowledge. In Table \ref{tb2}, \emph{AMC+Fixed} also achieves little size reduction on the model required a short decision trajectory. The incompleteness of the state setting for the AMC may explain the performance decline and the similar value of the compression ratio of \emph{AMC+Fixed}. Because the AMC using FLOPs(floating-point operations per second) reduction as an approximation of size reduction, it can not perform control of compression as precise as the AJPQ and end in local optimal strategy. Meanwhile, the AJPQ achieves 0.744 FLOPs reduction with 0.001 performance increase on SkyNet.

\textbf{Layer-wise Deep Quantization:} As demonstrated in Table \ref{tb1}, our \emph{AJPQ (layer-wise)} achieves a 1.8\% higher top-5 accuracy with a 0.012 larger size reduction
than the \emph{AMC+HAQ}. Taking advantage of the one-step compression, AJPQ saves nearly half of the time consumption caused by the serial compression process shown in Figure \ref{fig1}. For SkyNet, although the HAQ achieves performance increase by the necessary exploration for 300 searching episodes, the best result of \emph{AMC+HAQ} is not the optimal compression strategy. Based on global decisions, the AJPQ reduces more storage and computation than \emph{AMC+HAQ} with 0.188 compressed ratio and 0.931 FLOPs ratio.
In AJPQ, we implement different layer selection rules to adjust the quantization plan: the "large layer first" in ours and the "deep layer first" rule
in the HAQ. With a 0.09 higher top-1 accuracy and a 0.004 larger size reduction, our layer selection rule is more efficient than the HAQ's as shown in Table \ref{tb4}.
\begin{table*}[t]
\centering
\begin{tabular}{lccccc}
\toprule
Methods                  & Accuracy       & Accuracy       & Compressed          & FLOP                & Searching \\
$ $                      &(top-1)          &(top-5)         &ratio (bit)                     &ratio             &  episodes                  \\
\midrule
AJPQ (Deep layer first)  & 66.190          & 87.00        & 0.199                       & 0.808                 & 500                \\
AJPQ (Large layer first)   & \textbf{66.280}          & \textbf{87.00}        & \textbf{0.195}                       & \textbf{0.808}                 & \textbf{500}          \\\bottomrule
\end{tabular}
\caption{Performance achieved by AJPQ under different layer selection strategies.}
\label{tb4}
\end{table*}

 \begin{figure}[htp]
  \centering
  \includegraphics[width= 1.1\linewidth ]{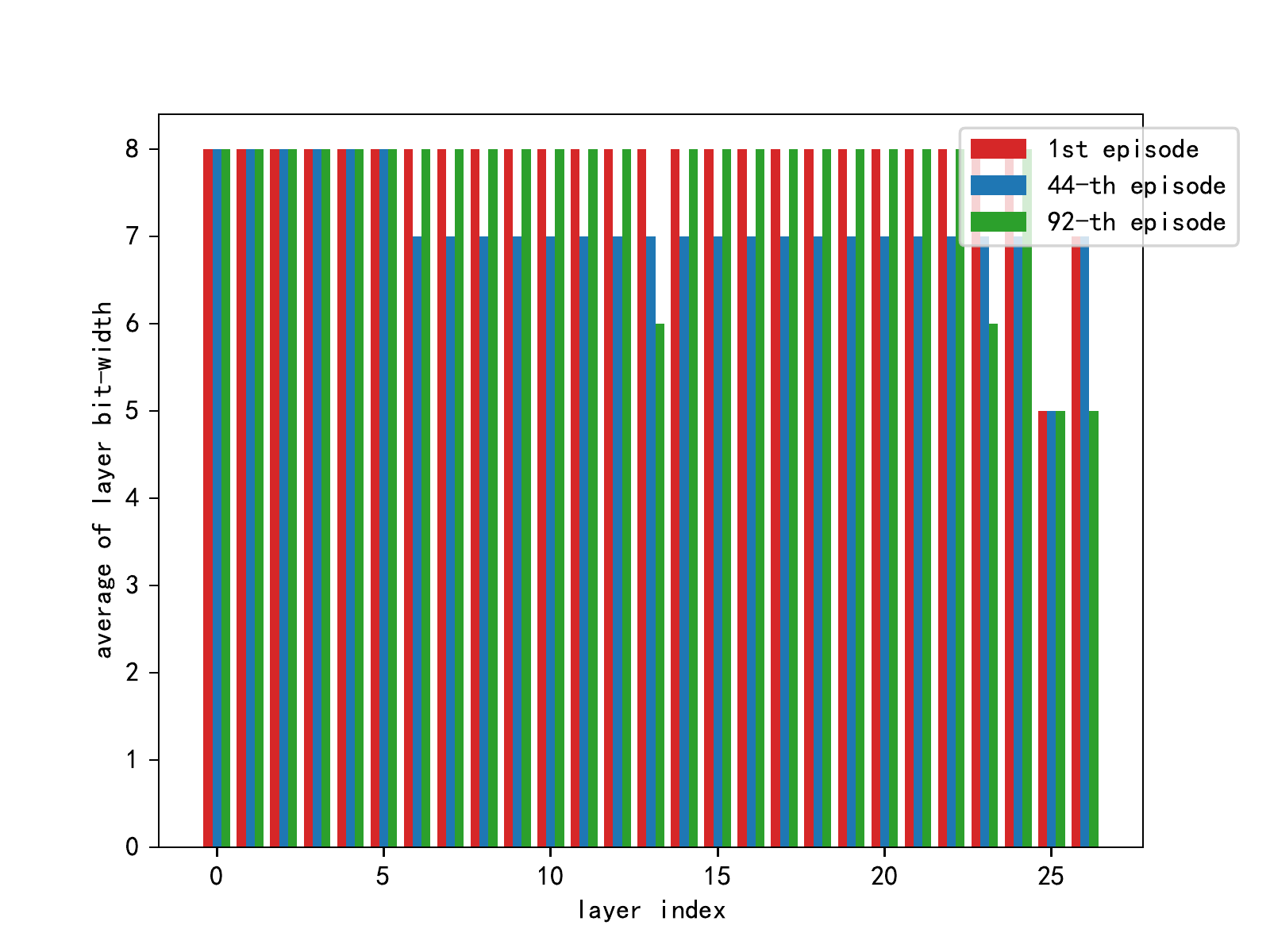}
  \caption{Quantization Plan Changes of MobileNet During Searching Under Compression Ratio = 0.2}
  \label{fig4}
 \end{figure}
\begin{figure}[htp]
  \centering
  \includegraphics[width= 1.1\linewidth ]{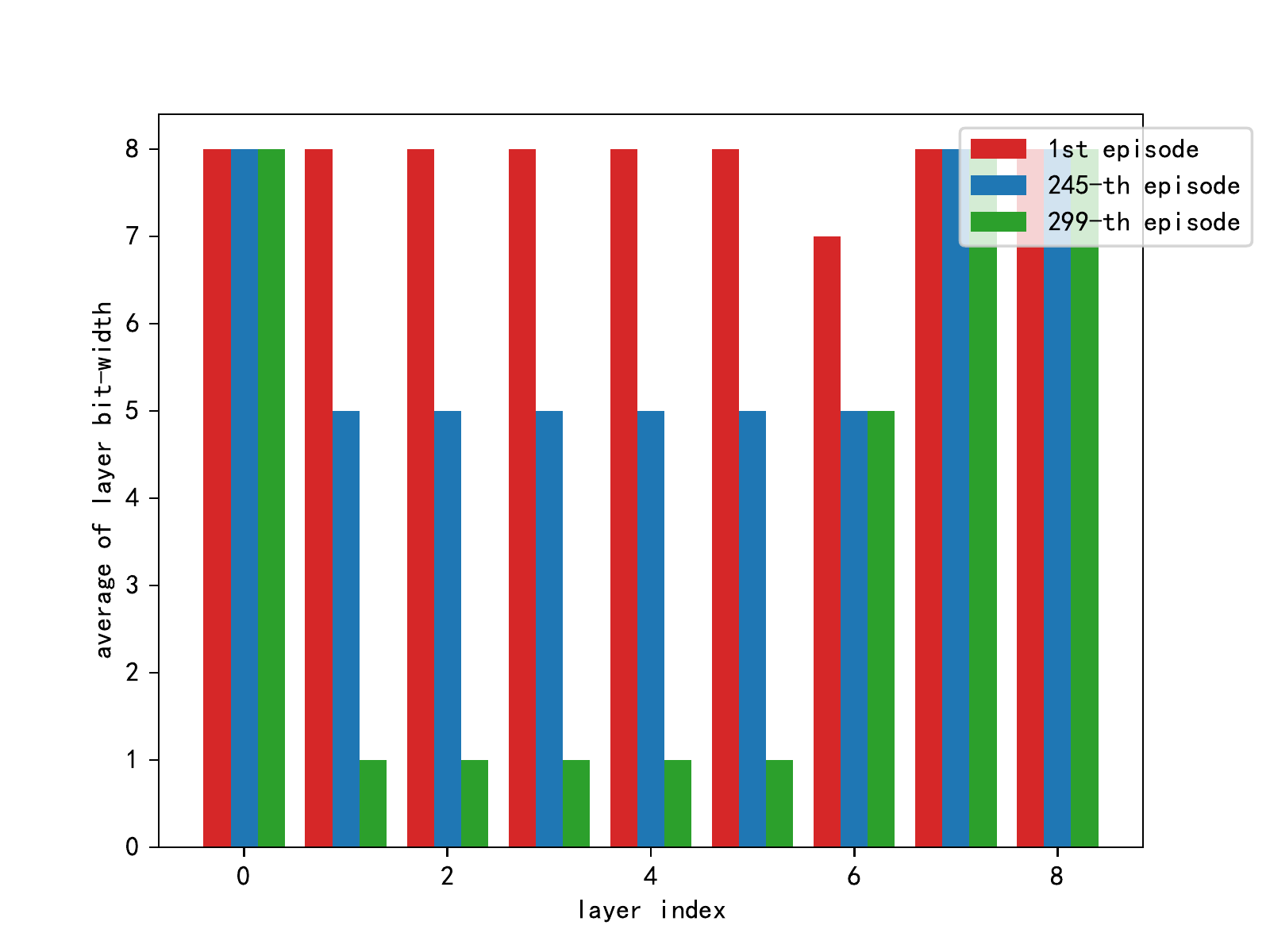}
  \caption{Quantization Plan Changes of SkyNet During Searching Under Compression Ratio = 0.2 }
  \label{fig5}
 \end{figure}

\textbf{Quantization Plan Changes During Searching:} To demonstrate how AJPQ performs to some specific compression requirements, we set the compression ratio to 0.2, and performance decline should less or equal with 0 for the classification task and detection task. As shown in Figure \ref{fig4}, a few layers perform quantization at the beginning. To meet the compression requirement, AJPQ tends to decrease bit-width to store weights for each layer. The best quantization plan performs more quantization in the last two layers than the other layers before. Because the fully connected(FC) layers in MobileNet contains more parameters than convolution layers. As shown in Figure \ref{fig5}, AJPQ tends to preserve channels rather than simply pruning. Compression is less performed on the last few layers than the former layers during the searching. Because Skynet uses the final convolution layer as the detector, the last few layers may be responsible for delivering the extracted features.
 
 \begin{figure}[htp]
  \centering
  \includegraphics[width= 1.15\linewidth ]{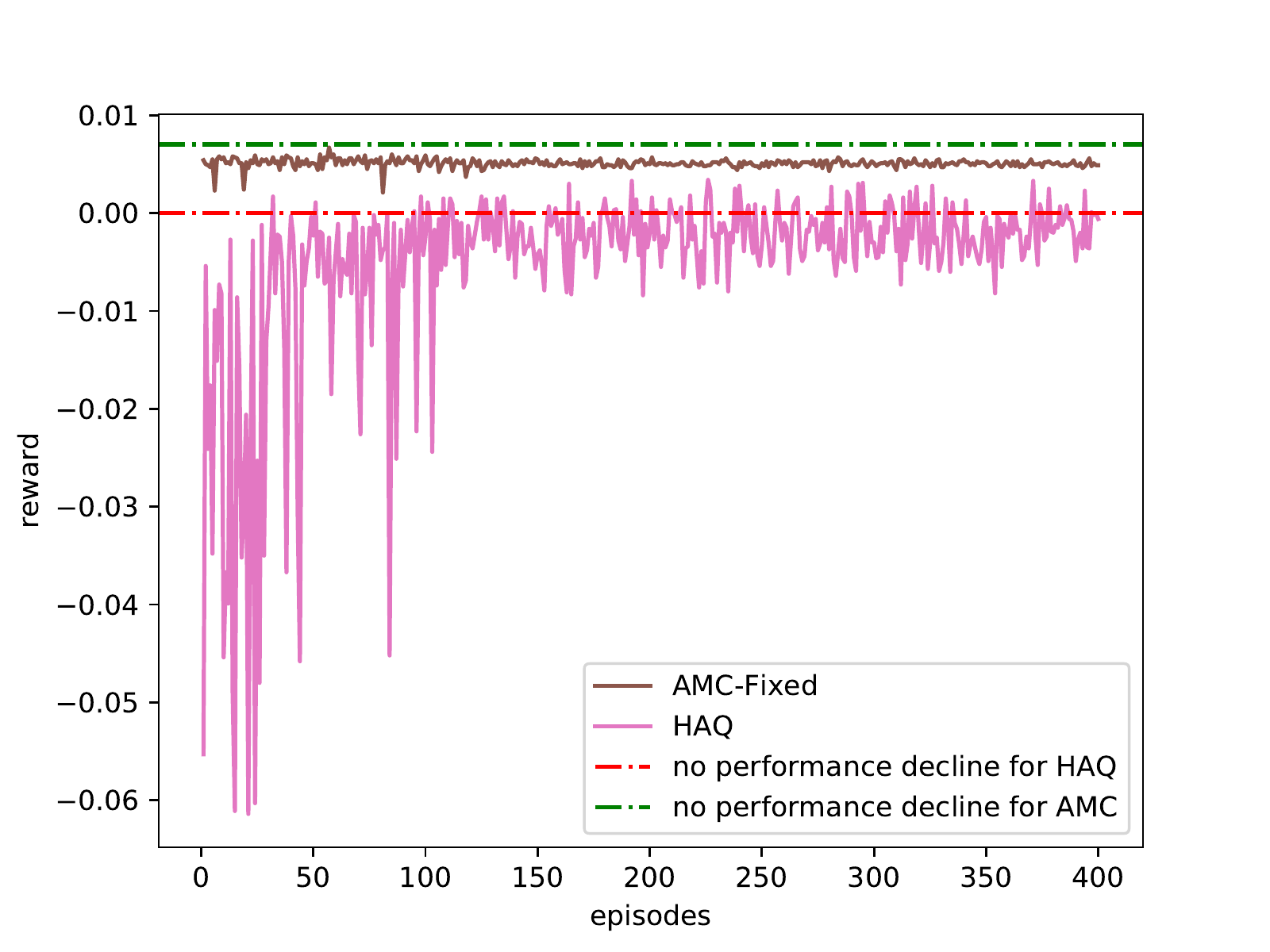}
  \caption{Rewards During Searching of AMC and HAQ  Under Compression Ratio=0.2}
  \label{fig6}
 \end{figure}
 
\begin{figure}[htp]
  \centering
  \includegraphics[width= 0.8\linewidth ]{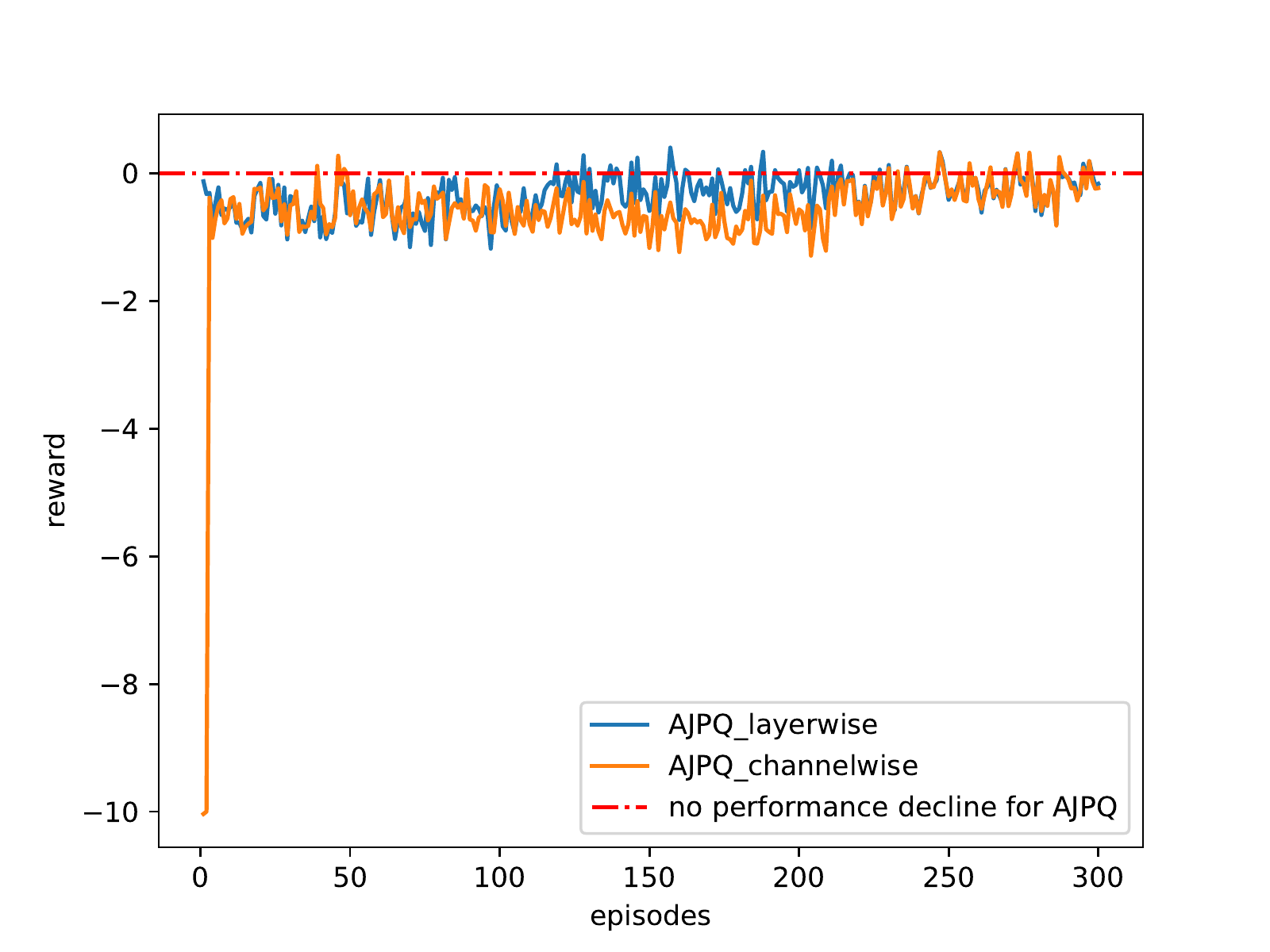}
  \caption{Rewards During Searching of AJPQ Under Compression Ratio=0.2}
  \label{fig7}
 \end{figure}

\textbf{Effectiveness evaluation of reward functions:} The reward function is defined as $reward = acc(\mathcal{N}') \times 0.01$ in AMC and $reward = (acc(\mathcal{N}') -acc(\mathcal{N}))\times 0.1$ in HAQ , where $acc(\mathcal{N}')$ is the performance of the compressed model and $acc(\mathcal{N})$ is the performance of the original model. The performance of those reward functions is shown in Figure \ref{fig6}. During the search, the AMC's rewards are hovering below the target reward. Meanwhile, the HAQ's rewards endure great fluctuation during the first 100 episodes. The AMC reward function gives all compression plans positive feed-backs without any discrimination. Thus, it can not efficiently explore. Meanwhile, HAQ's reward function can distinguish the strategies that can maintain performance. However, the exploration noise of the reinforcement learning agent is not the answer to the occurrences of the vibration. The small multiplier in the HAQ's reward function may weaken the punishment effect of negative reward. Our reward function can filter the unqualified searching strategies. As shown in Figure \ref{fig7}, the quantization plans violating the size reduction requirement are punished with a negative reward. Thus,the further explorations can focus on retaining performance.

\section{Conclusion}\label{sec:con}
In this paper, we proposed AJPQ (Automated Model Compression by Jointly Applied Pruning and Quantization), which proved that collaborating quantization and pruning could obtain a better trade-off between the performance holding and computation reduction. After providing a new viewpoint of pruning and quantization, we proposed a hierarchical manner to perform the quantization with mixed precision. A variety of experiments demonstrated the effectiveness of the proposed joint compression. In the future, the improvement in compression plan scheduling may be achieved by hierarchical reinforcement learning \cite{jung2019n}.



\bibliography{AJPQ}
\end{document}